\documentclass[10pt,twocolumn,letterpaper]{article}

\usepackage{cvpr}
\usepackage{times}
\usepackage{epsfig}
\usepackage{graphicx}
\usepackage{amsmath}
\usepackage{amssymb}
\usepackage{bm}
\usepackage{mathtools}
\usepackage[inline]{enumitem}
\usepackage[normalem]{ulem}
\useunder{\uline}{\ul}{}
\DeclarePairedDelimiter\norm{\lVert}{\rVert}%

\newcommand{\Eqq}[2]{
  \begin{equation}\label{#2}
  \begin{split}
  #1
  \end{split}
  \end{equation}
}

\usepackage[pagebackref=true,breaklinks=true,letterpaper=true,colorlinks,bookmarks=false]{hyperref}

\cvprfinalcopy % *** Uncomment this line for the final submission

\def\cvprPaperID{} % *** Enter the CVPR Paper ID here

\ifcvprfinal\fi
\begin{document}

%%%%%%%%% TITLE
\title{Deception Detection by 2D-to-3D Face Reconstruction from Videos}
\author{\rule[0pt]{12pt}{0pt} Minh L. Ng\^{o}$^{1,4}$
\and Burak Mandira$^{2}$
\and Selim Fırat Yılmaz$^{2}$
\and Ward Heij$^{1,3}$
\and Sezer Karaoglu$^{1,4}$
\and Henri Bouma$^{3}$
\and Hamdi Dibeklioglu$^{2}$
\and Theo Gevers$^{1,4}$
\and $^1$ University of Amsterdam
\and $^2$ Bilkent University
\and $^3$ TNO
\and $^4$ 3DUniversum
\and {\rule[0pt]{-15pt}{0pt} \tt\small \{l.m.ngo, th.gevers\}@uva.nl, \{burak.mandira,firat.yilmaz\}@ug.bilkent.edu.tr}
\and {\tt\small \{ward.heij, henri.bouma\}@tno.nl, s.karaoglu@3duniversum.com, dibeklioglu@cs.bilkent.edu.tr}
}

\maketitle
%\thispagestyle{empty}

%%%%%%%%% ABSTRACT
\begin{abstract}
Lies and deception are common phenomena in society, both in our private and professional lives. However,  humans are notoriously bad at accurate deception detection. Based on the literature, human accuracy of distinguishing between lies and truthful statements is $54\%$ on average, in other words it is slightly better than a random guess.
While people do not much care about this issue, in high-stakes situations such as interrogations for series crimes and for evaluating the testimonies in court cases, accurate deception detection methods are highly desirable. To achieve a reliable, covert, and non-invasive deception detection, we propose a novel method that jointly extracts reliable low- and high-level facial features namely, 3D facial geometry, skin reflectance, expression, head pose, and scene illumination in a video sequence. Then these features are modeled using a Recurrent Neural Network to learn temporal characteristics of deceptive and honest behavior.
We evaluate the proposed method on the Real-Life Trial (RLT) dataset that contains high-stake deceptive and honest videos recorded in courtrooms. Our results show that the proposed method (with an accuracy of $72.8\%$) improves the state of the art as well as outperforming the use of manually coded facial attributes ($67.6\%$) in deception detection.
\end{abstract}

%%%%%%%%% BODY TEXT
\section{Introduction}
Deceptive behavior is frequently displayed in daily life, yet, recognition of such behavior or lies is not an easy task for humans. On average, people are able to correctly classify only $47\%$ of lies and $61\%$ of truthful statements~\cite{bond2006accuracy}.
Therefore, reliable methods for deception detection is an
important need specifically for high-stakes situations such as
court cases, and suspect/witness interrogations for further
investigation. However, the ubiquitous polygraph, the most
commonly known lie detection mechanism, has been shown to be
unreliable~\cite{fiedler2002current}.

Invasive approaches such as PET (positron emission tomography)
and fMRI (function magnetic resonance imaging) based methods
perform better but they are neither fully reliable nor
practical in many situations where compactness or
portability is required.
Besides, the invasive nature of such mechanisms leaves them to
be easily tricked by skilled
deceivers~\cite{fiedler2002current}. Hence, deception detection
requires non-invasive and covert methods for accurate
detection. The difficulty in non-invasive deception detection
lies in the weakness of external cues, since a large volume of
work indicates that lies are barely evident in
behaviour~\cite{hartwig2014lie}.

Recent developments in computer vision, along with the availability of deceptive behavior videos, have increased the research interest on deceit detection from visual patterns.
The driving mechanism behind this ambition is the (subconscious) leakage of behavioral cues to deception~\cite{hartwig2014lie}. These cues are often weak,
very fast or subjective, making them hard to interpret by humans.
Recent studies on automated deception detection~\cite{Morales2017} exploits different behavioral modalities such as facial actions/expressions, head pose/movement, gaze, hand gestures, and even vocal features in the analysis~\cite{Morales2017, abouelenien2014deception}. In contrast, our work focuses solely on facial cues (including head pose/movement), yet providing a better accuracy.   

High-level visual features used in the literature~\cite{Morales2017} such as facial action units, are prone to errors due to challenging environmental conditions (i.e. illumination, view point, occlusion etc.). Thus, such features can introduce significant amount of noise in the analysis. In this paper, to cope with such issues, we propose to exploit face reconstruction to obtain an effective low level representation for a more reliable deceit detection. Face reconstruction aims at decomposing a face image into its components such as 3D facial geometry, expression, skin reflectance, head pose, and illumination parameters, which are expected to carry important information for deceit detection. While the illumination parameters sound like unrelated to be used in this task, in combination with geometry it reveals subtle changes in expression-related skin deformations. Furthermore, prediction of these parameters, in our method, are constrained by an image formation model that relies on joint parametric modeling of facial cues, head pose, and illumination. Therefore, it minimizes the possible negative influences of varying environmental conditions.
Once such components are extracted, they are fed to a Recurrent Neural Network to model temporal characteristics of deceptive and honest behavior in videos.

Although, a successful decomposition has been a backbone for many face-related computer vision tasks (e.g. face recognition, emotional expression recognition, head pose estimation, etc.), this work is the first one that exploits face reconstruction for deceit detection. Furthermore, we propose a fully unsupervised end-to-end deep architecture for face reconstruction (including 3D facial geometry, expression, reflectance, head pose and illumination) from videos. Our results show that the proposed novel method for deception detection improves the state of the art, as well as outperforming the use of manually annotated facial attributes (e.g. facial actions/expressions, gaze, and head movement) for this task.

\section{Related Work}

\subsection{Deception Detection} 
At the basis of deception detection through nonverbal cues stands the leakage hypothesis, which states that --if the stakes of a lie are high enough-- involuntary, subconscious cues of deceit will emerge from a liar~\cite{hartwig2014lie}. One can divide observable cues in physiological cues, body language cues and facial cues. One of the problems about intangible constructs such as deceit is that these cues range from highly objective ones (vocal pitch) to highly subjective measurements (facial pleasantness). Hence, this section aims to provide an overview of objective, non-verbal cues that are relevant to the scope of using visual features for deception detection.

Concerning facial cues, a multitude of signals have been identified to correlate with deceit, such as lip pressing~\cite{burgoon2017social}, smiling and pupil dilation and facial rigidity~\cite{pentland2017video}. However, the studies often find contradictory results~\cite{bouma2016measuring}. In addition, performance is highly dependent on the data used for training and validation, with some datasets being noticeably easier than others~\cite{wu2017you}. Secondly, the circumstances under which the lies were elicited are influential: multiple studies indicate that deceptive cues increase in magnitude with increased cognitive load~\cite{vrij2017cognitive}. Hence, the final application and training data should have comparable cognitive load during data recording.

Micro-expressions pose another viable source of information~\cite{yan2018measuring}, even though other studies have shown that only a small amount of people exhibit micro-expressions when lying~\cite{desjardins2015reading}. Facial action units (AUs) are also found to be informative for deceit detection~\cite{Morales2017}. 

One of the most recent methods on automated deceit detection is proposed by Morales~\etal~\cite{Morales2017}. This method fuses information from audio-visual modalities, where visual features in the form of 408 cues, including gaze, orientation and FACS information, are extracted using OpenFace~\cite{baltruvsaitis2016openface} and later fused with verbal and acoustic features. Fusion occurs through concatenation of statistical functional vectors, after which random forests and decision trees are used for deception classification. Differently, \cite{Perez-Rosas:2015:DDU:2818346.2820758} presents a baseline method for their introduced Real-Life Trial dataset, which models manually coded visual features such as expression, head movement, and hand gestures together with speech transcriptions using random forests and decision trees.

\subsection{Monocular Face Reconstruction}
Decomposition of image components requires inverting the complex real-world image formation process. The reconstruction by inverting image formation is an ill-posed problem because infinite number of combinations can produce the same 2D image~\cite{blanz1999morphable}. In general, we can categorize face reconstruction methods into two groups, namely, iterative~\cite{blanz1999morphable,thies2015real,garrido2013reconstructing,thies2016face2face} and deep learning based~\cite{tewari17MoFA}. Iterative approaches try to optimize parameters by minimizing the error between projected (reconstructed face) and original image in an iterative (analysis-by-synthesis) manner~\cite{blanz1999morphable}. The energy functions are mostly non-convex. Good fitting can only be obtained by close initialization to the global optimum, which is only possible with some level of control during image capture. Since  these approaches are computationally expensive therefore are not preferred in this paper. 

Deep learning based methods to reconstruct face from a single monocular image typically uses either data augmentation techniques to regress prediction to be close to the ground truth~\cite{InverseFaceNet, Genova_2018_CVPR} or applies the similar analysis-by-synthesis approach to train the neural network using a physically plausible image formation model~\cite{tewari17MoFA, Genova_2018_CVPR}. These methods produce sufficient reconstruction quality for certain tasks, however, they sacrifice details in order to be tractable for challenging, unconstrained images. Since such methods cannot avoid expression information to be leaked in 3D facial geometry, it is likely that there is an information loss while capturing expression. To reliably capture facial movements, the separation of 3D facial geometry and expression components are quite important.

Some works have been proposed to overcome such issues by using RGB videos instead of single monocular images~\cite{thies2015real,garrido2013reconstructing,thies2016face2face}. However, these works are based on iterative approach. Convolutional Neural Network (CNN) architectures are rarely explored for video-based dense real-time face reconstruction. In this paper, we present a novel identity-aware, dense and real-time face reconstruction CNN pipeline which receives RGB videos as input. Unlike previous monocular reconstruction methods, our method extracts identity related parameters (i.e. 3D facial geometry and reflectance) for a full video sequence whereas temporally dependent parameters (i.e. expression and illumination) for every single frame. The proposed method prohibits leakages of expression parameters to 3D facial geometry by temporal constraints which improves the preciseness of facial expression capture. Furthermore, using a Recurrent Neural Network (RNN), we temporally constrain the expression so that we preserve the consistency between expression through full video.

\section{Methodology}

\subsection{Network architecture}

Convolutional Neural Network is used to predict intrinsic inverse rendering parameters $\mathcal{P}\in\mathbb{R}^{257}$ (code vector) from a set of RGB face images $\{\mathbf{I}_i\} \in \mathbb{R}^{W\times H\times 3}$, from which a reconstructed image can be recovered:

\Eqq{
	\mathcal{P} &= \{ \bm{\alpha}, \bm{\beta}, \bm{\delta}, \bm{\gamma}, \bm{\omega}, \bm{t} \},
}{eq:code_vector}

where $\bm{\alpha} = \{\alpha_i\},~\bm{\beta} = \{\beta_i\}\in\mathbb{R}^{80}$ and $\bm{\delta}=\{\delta_i\}\in\mathbb{R}^{64}$ are parameters corresponded to 3D face geometry, reflectance and expression; $\bm{\gamma}\in\mathbb{R}^{9\times 3}$ represents scene illumination parameters; $\bm{\omega}\in \mathbb{SO}(3)$ and $\bm{t}\in \mathbb{R}^3$ represent rotation and translation parameters.

\begin{figure*}
\begin{center}
\includegraphics[width=\linewidth]{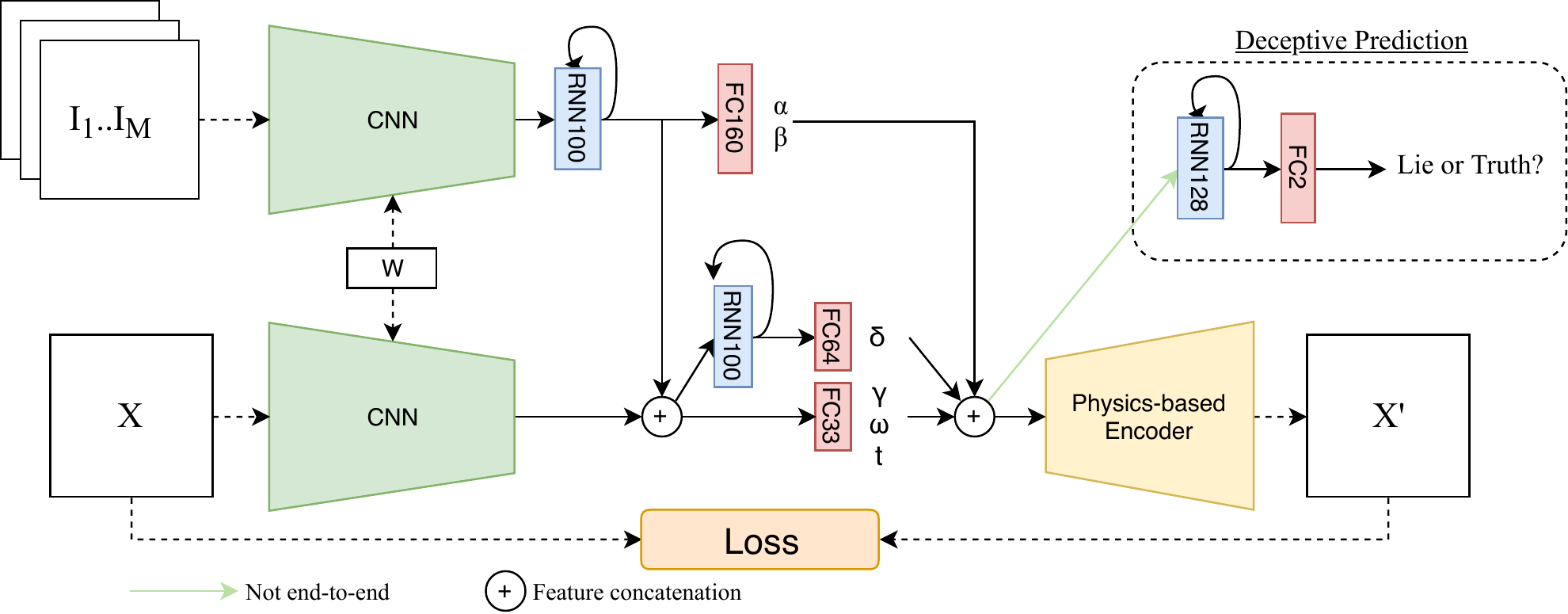}
\end{center}
   \caption{Architecture overview. Our pipeline consists of 2 AlexNet with shared weights. One for extracting features for identity parameters, another to target frame-wise parameters. Backbones are followed by recurrent layer units and fully connected layer to predict semantic code vector. We train our pipeline using physics-based encoder constraining code vector to be able to produce the close reconstruction to an input image. Predicted code vector is used during the testing time for deceptive prediction pipeline which is trained separately.}
\label{fig:architecture_overview}
\end{figure*}

Figure~\ref{fig:architecture_overview} shows an overview of our face reconstruction architecture. Our model consists of two AlexNet~\cite{Krizhevsky:2012:ICD:2999134.2999257} backbones with shared weights, one (Identity CNN) to extract person identity 3D facial geometry and reflectance features related to $\bm{\alpha}, \bm{\beta}$ from a collection of images $\mathbf{I}_1 .. \mathbf{I}_M$ and another (Framewise CNN) to extract frame-dependent facial features $\bm{\delta}, \bm{\gamma}, \bm{\omega}, \bm{t}$ from particular frame $\mathbf{X}$. For our purpose we are using all layers of AlexNet except the last FC8 layer. Those features are fused using recurrent units with 100 hidden parameters and fully connected layers without non-linearity to predict single set of identity parameters $\bm{\alpha}$, $\bm{\beta}$ given a set of cropped face $\mathbf{I}_1$..$\mathbf{I}_M$, expression parameters $\bm{\delta}$ conditioned on the previous temporal state, illumination, rotation and translation parameters.

Identity CNN is followed by recurrent unit of 100 hidden parameters and fully connected layer without non-linearity to produce identity parameters $\bm{\alpha}$, $\bm{\beta}$. Features from a recurrent unit is concatenated with Framewise CNN. This representation is followed by another recurrent unit of 100 hidden parameters and fully connected layer to produce blend shape parameters, and just fully connected layer without recurrent unit for other parameters.

\subsection{Physics-based image formation}

\textbf{3D facial geometry and reflectance}. We parametrize 3D face geometry using a multi-linear PCA model~\cite{DBLP:journals/corr/abs-1709-08398} separately for neutral face and face expression (Eq.~\ref{eq:geom}). 3D face geometry is represented as a point cloud $\mathbf{X} = (\mathbf{x}, \mathbf{y},\mathbf{z})^T\in\mathbb{R}^{N\times3}$ in the Euclidean space.

\Eqq{
	\mathbf{X} &= \mathbf{A_{id}} + \sum_{i=1}^{80}\alpha_i \sigma_{id,i} \mathbf{P_{id,i}} + \mathbf{A_{exp}} + \sum_{i=1}^{64}\delta_i \sigma_{exp,i}\mathbf{P_{exp,i}}
}{eq:geom}

where we denote $\mathbf{A_{id}}$, $\mathbf{A_{exp}}\in \mathbb{R}^{N \times 3}$ as an average neutral face and an average expression geometries, $\mathbf{P_{id,i}}$, $\mathbf{P_{exp,i}}\in\mathbb{R}^{N\times 3}$ as their principal components sorted by standard deviations $\sigma_{id,i}$, $\sigma_{exp,i}\in\mathbb{R}$ respectively. Face reflectance is modelled using a separate PCA model: 

\Eqq{
\mathbf{B} &= \mathbf{A_{tex}} + \sum_{i=1}^{80}\beta_i \sigma_{tex,i} \mathbf{P_{tex,i}}
}{eq:reflect}

where $\mathbf{A_{tex}}\in\mathbb{R}^{N\times 3}$ is an average face reflectance, $\mathbf{P_{tex,i}}\in\mathbb{R}^{N\times 3}$ are principal components sorted by standard deviations $\sigma_{tex,i}\in\mathbb{R}$.

\textbf{Face transformation}. We model face movement in the scene using 6DoF transformation $\mathbf{T}$. Rotation matrix $\mathbf{R}(\bm{w}): \mathbb{R}^3\to \mathbb{R}^{3\times 3}$ is represented in $\bm{\omega}\in\mathbb{R}^3 \in \mathbb{SO}(3)$, and translation $\bm{t}\in\mathbb{R}^3$ is separate in each x, y, z directions.

\textbf{Illumination model}. Illumination changes are modelled using first 3 bands of spherical harmonics basis function $\mathbf{H}_{j}$ assuming face is a Lambertian surface \cite{thies2016face}. Intensity of the i-th vertex $c_i$  is defined as a product of vertex reflectance $b_{i}$ and a shading components.

\Eqq{
c_i &= b_i\sum_{j=1}^{3^2}\gamma_{j}\mathbf{H}_{j}(\mathbf{R}(\bm{\omega})\mathbf{n}_i), i \in 1..N,
}{eq:shading}

where $\mathbf{n}_i$ is a vertex normal of the i-th vertex. We define illumination parameters $\gamma_j$ separately for each R, G, B channels and thefore have 27 parameters in total. Vertex normal is estimated based on 1-ring triangle neighbours. Triangle topology is known from the face morphable model.

\textbf{Projection model}. An obtained 3D point cloud $\mathbf{X}$ is mapped into a 2D plane by applying a rigid transformation $\mathbf{T}$ and  perspective transformation $\mathbf{\Pi}$ which is modelled as product of projection $\mathbf{V}\in\mathbb{R}^{4\times 4}$ and viewport $\mathbf{P}\in\mathbb{R}^{4\times 4}$ matrices:

\Eqq{
\begin{bmatrix}
  \hat{x} \\
  \hat{y} \\
  \hat{z} \\
  \hat{d} \\
\end{bmatrix} &= \underbrace{\begin{bmatrix}
	\mathbf{V}
\end{bmatrix} \times \begin{bmatrix}
	\mathbf{P}
\end{bmatrix}}_{\mathclap{\mathbf{\Pi}}} \times \underbrace{\begin{bmatrix}
  \mathbf{R}(\bm{\omega}) & \mathbf{t} \\
  \mathbf{0} & 1
\end{bmatrix}}_{\mathclap{\mathbf{T}}} \times \begin{bmatrix}
  x \\
  y \\
  z \\
  1
\end{bmatrix}.
}{eq:proj}

$\hat{u}, \hat{v}$ coordinates and depth can be obtained by division by the homogeneous coordinate $\hat{d}$. Focal length is assumed to be fixed and principal points to be in the middle of the projection screen. 

\subsection{Training loss}

We employ the energy minimization strategy of Tewari \etal \cite{tewari17MoFA}. In total our loss contains of 3 main components: landmark loss $E_{land}$, vertex-wise photometric loss $E_{vert}$ and regularization term $E_{reg}$ (Eq.~\ref{eq:loss}).

\Eqq{
\mathcal{L} &= w_{land} E_{land} + w_{vert} E_{vert} + E_{reg}
}{eq:loss}

\textbf{Landmark loss}. $L_2$ difference between landmark projections $p$ from a predicted 3D face model and ground truth landmark $l_j$ are used. In total, we use $|\mathcal{F}| = 48$ landmarks for optimization covering eyebrows, eye corners, nose, mouth and chin regions.

\Eqq{E_{land} &= \dfrac{1}{|\mathcal{F}|}\sum_{j\in\mathcal{F}} \norm{p_{k_j} - l_j}^2_2}{eq:lm_loss}

where we define $k_j$ as an annotated vertex index of the j-th landmark on the 3D model.

\textbf{Vertex-wise photometric loss}. We define photometric loss as a $L_{2,1}$ difference \cite{Ding:2006:RPR:1143844.1143880} between vertex intensity color and its corresponded color from the original image. To find an intensity color on image space we perform bilinear interpolation.

\Eqq{E_{vert} &= \frac{1}{|\mathcal{V}|}\sum_{i\in\mathcal{V}} \norm{\mathbf{c}_i - \mathbf{X}_{\hat{u}_i, \hat{v}_i}}_2,}{eq:full_loss}

We filter out vertices which contributes to the photometric loss based on normal direction, $|\mathcal{V}|$ is an amount of those vertices.

\textbf{Statistical regularization}. We use Tikhonov regularization~\cite{thies2016face} to enforce predicted parameters to be in the plausible range.

\Eqq{
E_{reg} &= w_{\alpha} \sum_{i=1}^{80} \alpha^2_i + w_{\beta} \sum_{i=1}^{80} \beta^2_i + w_{\delta} \sum_{i=1}^{64} \delta^2_i
}{eq:reg}

\subsection{Modeling Deceptive Behavior}
Once the facial representation is obtained, we classify videos as deceptive or honest. We employ recurrent neural network (RNN) to capture temporal relations between facial representation vectors of frames $\mathcal{P} \in \mathbb{R}^{257}$ for each video. We use the loss function
\Eqq{
\mathcal{J} &= \sum_{j=0}^{1} (y_{i, actual} - y_{i, pred}^j)^2
}{eq:cl_loss}
where $y_{i, actual}$ is the label of the video i, and $y_{i, pred}^j$ is the predicted probability of video i for class j. Deceptive and honest behaviors correspond to 1 and 0, respectively.

We employ single layer RNN of 128 units followed by 2 output neurons with sigmoid activation function. Each output neuron corresponds to a class, deception or truth. At evaluation stage, the class that corresponds to the output neuron with maximum probability is the final prediction.

\begin{figure*}[t!]
    \centering
    \includegraphics[width=\textwidth]{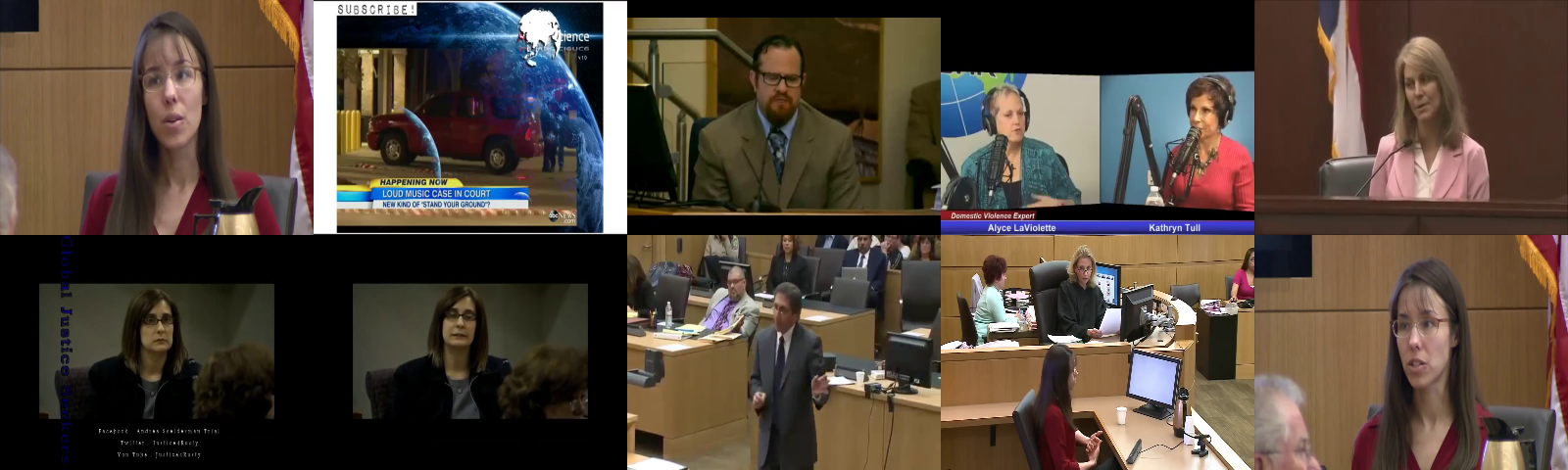}
    \caption{Sample video frames from the RLT dataset. Dataset contains video of trials under different lighting conditions, pose, with multiple people in the scene. Some of videos are heavily occluded and doesn't contain visible facial features.}
    \label{fig:rlt}
\end{figure*}

\section{Implementation details}

We train our 2D-to-3D face reconstruction network for 200K iterations on 300VW \cite{DBLP:journals/ijcv/ChrysosASAZ18} and CelebA datasets \cite{liu2015faceattributes} using a batch size of 5 and Adam optimizer \cite{DBLP:journals/corr/KingmaB14} with learning rate of $10^{-5}$. Loss weights are set to be $w_{vert} = 1.92$, $w_{land} = 0.0019$, $w_{\alpha} = 2.9\times 10^{-5}$, $w_{\beta} = 4.93\times 10^{-8}$, $w_{\delta} = 2.32\times 10^{-5}$.

300VW contains video sequences with annotated 68 landmarks for each frame. We crop faces based on a bounding box on ground truth landmarks with 10\% expansion. We process CelebA using dlib \cite{dlib09} for face detection and FAN \cite{bulat2017far} for landmark detection. In total we have collected 94K images from 300VW coming from 49 videos and 200K images from CelebA.

For each video sequence of 300VW we randomly select 3 crop faces in random order as an input for the Identity-CNN. We randomly sample a sequence of 3 crop faces with a random step size from 1 to 5 frames as an input for the Framewise-CNN. For CelebA we assume that we have a 1-frame video sequence for each image. Images are randomly flipped to augment the dataset size. We train the model alternating CelebA and 300VW batches.

AlexNet backbones are initialized using a pretrained model on ImageNet. We add additional offset to the 0-th band SH coefficient and z-translation to make sure initial 3D face model has a plausible initial illumination and is centered in the middle of the screen.

Basel Face Model 2017 \cite{DBLP:journals/corr/abs-1709-08398} is used for 3D face geometry, reflectance and expression. We take first 80 principal components for $\bm{\alpha}$ and $\bm{\beta}$ and 64 for $\bm{\delta}$. We implement $\mathbb{SO}(3)$ gradients in the compact form based on the work of Gallego \etal\cite{DBLP:journals/corr/GallegoY13}. Our implementation is written in Tensorflow \cite{tensorflow2015-whitepaper}. We ran our experiments on NVIDIA GTX 1080.

We train our deception modeling RNN for 10 epochs on Real-Life Trial (RLT) dataset using a batch size of 16 and Adam optimizer \cite{DBLP:journals/corr/KingmaB14} with learning rate $10^{-3}$.

\section{Dataset}
In this study, we employ the Real-Life Trial dataset~\cite{Perez-Rosas:2015:DDU:2818346.2820758} which contains 121 videos from real-life high-stakes scenarios that are publicly available. See Fig.~\ref{fig:rlt} for visual samples from dataset. It has 61 deceptive and 60 truthful trial clips of 21 female and 35 male subjects whose ages vary between 16 and 60. The average duration of videos is about 28 seconds. When constructing the dataset, Perez-Rosas \etal~\cite{Perez-Rosas:2015:DDU:2818346.2820758} enforce some visual constrains for videos such as the defendant or witness and his or her face should be clearly identified during most of the footage as well as some vocal enforcements which are not relevant within our context. 
We discard 40 of these videos from the dataset due to technical errors: 1) failure in facial landmark detection using \cite{dlib09} ~ 2) some videos do not display the target subject and instead show something else such as courtroom while having the voice of target subject. Thereby, a subset of 81 videos (39 deceptive, 42 truthful) from this dataset is used which is constructed from 28 male and 53 female subjects.

\section{Experiments}

In this section, we explain all the experiments that are conducted in detail. We considered lie as \textit{positive} and truth as \textit{negative} throughout the experiments when calculating accuracy, precision and recall.

\subsection{Comparison with monocular 2D-to-3D methods}

Our pipeline constraints prediction of identity parameters $\bm{\alpha}$, $\bm{\beta}$ by making use of randomly selected multiple frames. In this experiment, we evaluate how sensitive the proposed face reconstruction is to the choice of frames for identity estimation. We compare our network accepting random single monocular image against to the work of Tewari \etal which does not perform any conditioning on identity. The evaluation is performed on the 300VW validation split. It contains 3 video subsets with different scene complexity. Set 1 contains 31 videos, set 2 contains 19 videos with difficult illumination conditions, set 3 contains 14 videos with occlusion, extreme pose and expressions. Results are reported in the Table~\ref{tab:std}. Results have shown that our method produces more consistent predictions for albedo and shape parameters in comparison to Tewari \etal. This shows that the proposed method avoid leakages between expression and geometry. Consequently, proposed method predicts more precise expression and geometry.

\begin{table}[h]
\begin{center}
\begin{tabular}{l|r|r|r|}
\cline{2-4}
      & Set 1        &  Set 2 & Set 3       \\ \hline
\multicolumn{1}{|l|}{Tewari \etal\cite{tewari17MoFA} ($\bm{\alpha}$ std)} & 0.240 & 0.191 & 0.299\\ 
\multicolumn{1}{|l|}{Tewari \etal\cite{tewari17MoFA} ($\bm{\beta}$ std)} & 0.299 & 0.159 & 0.174\\
\hline
\multicolumn{1}{|l|}{Ours ($\bm{\alpha}$ std)} & \textbf{0.051} &	\textbf{0.052}	 & \textbf{0.064}\\ 
\multicolumn{1}{|l|}{Ours ($\bm{\beta}$ std)} & \textbf{0.065} &	\textbf{0.052}	 & \textbf{0.082}\\
\hline
\end{tabular}
\end{center}
\label{tab:std}
\caption{Identity parameters ($\bm{\alpha}$, $\bm{\beta}$) variation comparison on 300VW validation sets. Our network is accepting single monocular image during the testing time. Lower - better.}
\end{table}

\subsection{Face Reconstruction Visual Results}

\begin{figure*}
\begin{center}
\includegraphics[width=\linewidth]{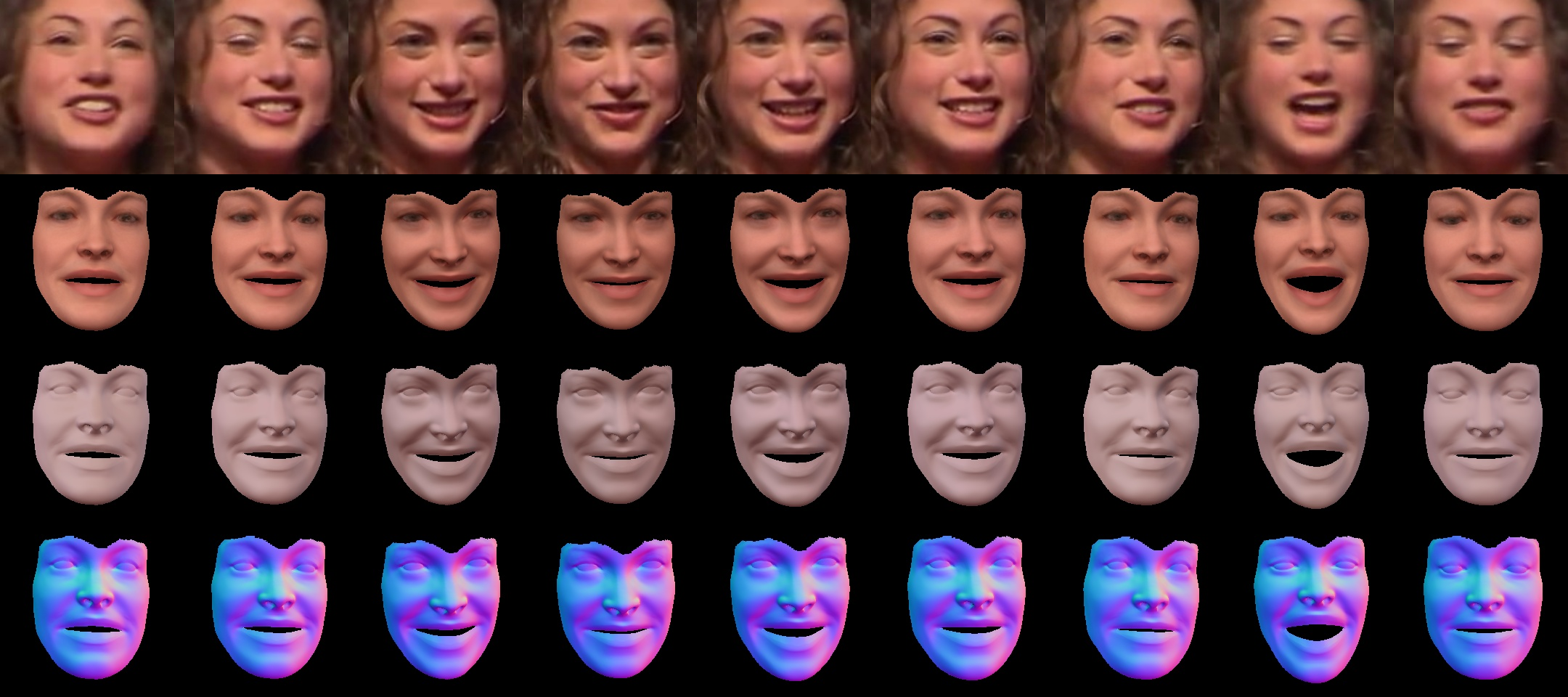}
\end{center}
   \caption{Visual prediction of our identity aware 2D-to-3D face reconstruction on the 300VW validation split. From top to bottom: original image, reconstructed result, shading, normals.}
\label{fig:visual}
\end{figure*}

We show visuals of our reconstruction pipeline in the Figure~\ref{fig:visual}. Our method successfully recover intrinsic properties of a face such as shading, normals and color intensity preserving facial identity over video frames.

\begin{table}[t!]
\begin{center}
\begin{tabular}{|c|c|c|c|c|}
\hline Gender & Accuracy & Precision & Recall & \# samples\\ \hline
\multicolumn{1}{|c|}{Male}   & 0.64     & 0.29      & 1.00  & 28  \\
\multicolumn{1}{|c|}{Female} & 0.77     & 0.87      & 0.77 & 53  \\ \hline
\end{tabular}
\end{center}
\caption{Gender specific deceit detection results.}
\end{table}

\subsection{Gender Effect}

In this experiment, we investigate the effect of gender to our results. The dataset does not provide gender annotation, therefore we annotate the dataset with gender information. Then dataset samples are grouped based on their gender to analyze results for each gender. The results are summarized in Table 2. High precision and recall values of females may suggest that feature extraction of females is more challenging and has high variation. However, this can also be related to the number of samples as we have female subjects almost as twice as males subjects.

\begin{table*}[t]
\begin{center}
\begin{tabular}{c|c|c|c|c|}
\hline
\multicolumn{1}{|c|}{Model}       & Feature        & Accuracy & Precision & Recall        \\ \hline
\multicolumn{1}{|c|}{Morales \etal\cite{Morales2017} (DT)*} & OpenFace features & 0.55 & 0.54 & 0.50\\
\multicolumn{1}{|c|}{Morales \etal\cite{Morales2017} (RF)*} & OpenFace features & 0.50 & 0.45 & 0.25\\
\multicolumn{1}{|c|}{Perez-Rosas \etal\cite{Perez-Rosas:2015:DDU:2818346.2820758} (DT)*}   & Hand-labeled features & 0.66 & 0.67 & 0.55\\ 
\multicolumn{1}{|c|}{Perez-Rosas \etal\cite{Perez-Rosas:2015:DDU:2818346.2820758} (RF)*}   & Hand-labeled features & 0.67 & \textbf{0.70} & 0.55\\
\multicolumn{1}{|c|}{Time-CNN \cite{zhao2017convolutional}}   & FRC & 0.69 & 0.65 & 0.77 \\ \hline
\multicolumn{1}{|c|}{Our RNN}   & FRC & \textbf{0.73} & 0.68 & \textbf{0.79}\\ \hline
\end{tabular}
\end{center}
\label{tab:exps}
\caption{Results of other models and the proposed deception modeling RNN on RLT dataset.}
\vspace{-0.4cm}
\begin{center}
\footnotesize *: only facial features are used
\end{center}
\end{table*}

\subsection{Comparison to Other Models}

First, we start our comparisons with reproducing the baseline models. The model of Morales \etal \cite{Morales2017} is tested with decision tree (DT) and random forest (RF) classifiers with default parameters as mentioned in their papers. Morales \etal use OpenFace~\cite{baltruvsaitis2016openface} to extract facial features in default output (i.e. basics, gaze, pose, 2D and 3D facial landmark locations, rigid and non-rigid shape parameters, action units) and apply some statistical functionals (i.e. maximum, minimum, mean, median, standard deviation, variation, kurtosis, skewness, 25\% percentile, 50\% percentile, and 75\% percentile) in order to create one feature vector per video.  

The model of Perez-Rosas \etal\cite{Perez-Rosas:2015:DDU:2818346.2820758}, which is the basis for Morales \etal~\cite{Morales2017}, is also implemented with decision tree (DT) and random forest (RF) classifiers with default parameters as mentioned in their papers. They use manually crafted features (i.e. smile, laughter, scowl, gaze, lips, openness and closeness of eyes and mouth, position of eyebrows like frowning and raising, head movements, hand trajectory and movements). Thus, the accuracy results of their work indeed show the performance of human annotators. Note that, since our system focuses only on facial features, we excluded hand-related features from their experiment setup to obtain comparable results.

Morales \etal\cite{Morales2017} mention 71.07\% and 73.55\% accuracy results for their visual model with DT and RF classifiers, respectively. However, they obtain these figures erroneously by applying \textit{leave-one-out} cross-validation which causes to subject overlaps between test and train dataset. Thereby, in our experiments for both \cite{Morales2017} and \cite{Perez-Rosas:2015:DDU:2818346.2820758}, we applied \textit{leave-one-person-out} (LOPO) cross-validation (where videos of a single subject are separated as test set and from the remaining videos five of them are randomly sampled as validation set and the remaining videos are taken as training set for each test fold) to reproduce their corrected accuracy results, given in Table~\ref{tab:exps}. Note that, in order to have balanced training dataset at the end, we randomly downsampled majority class in terms of quantity to have equal number of instances from each class. In Table 3, each DT and RF result is obtained by taking the average of 20 iterations.

As our third baseline, we experiment with convolutional neural network for time series classification (Time-CNN) model \cite{zhao2017convolutional}, as shown in Table 3. This method reveals time series patterns through 1D convolutions on temporal vector of each individual feature dimension. This model emphasizes the strength of our proposed deception modeling RNN as it constructs a strong baseline that uses the same features (i.e. face reconstruction components (FRC) that our proposed CNN face reconstruction network extracts) with the proposed RNN.

Last row of Table 3, recurrent neural network (RNN) model, shows the performance of our proposed deception detection method.

The results are summarized in Table~\ref{tab:exps}. The results show that the proposed RNN method improves the state of the art with 4\%. In addition, we also improve (6\%) \cite{Perez-Rosas:2015:DDU:2818346.2820758}, the method which uses manually annotated features. There might be leakage of behavioural cues to deception. This leakage may not be an obvious behaviour which may not necessarily annotated by the human observer. This may indicate that the proposed method captures even subtle behaviours for deceit detection. 

\section{Conclusion}

We have presented a novel method for deception detection based on reliable low- and high-level facial features obtained using 2D-to-3D face reconstruction technique. To be able to reconstruct faces (including 3D facial geometry, expression, reflectance, head pose and illumination) from videos, we propose a fully unsupervised end-to-end deep architecture. We show our method to produce consistent identity prediction in contrast to other deep learning methods which take only one monocular frame during the testing time. Our pipeline uses recurrent neural networks to learn temporal behaviour, works real time and shows state-of-the-art accuracy in the challenging Real-Life Trial (RLT) dataset. Our results show that the proposed method (with an accuracy of 72.8\%) improves the state of the art as well as outperforming the use of manually coded facial attributes (67.6\%) in deception detection.

{\small
\bibliographystyle{ieee}
\bibliography{main}
}

\end{document}